\documentclass{article}

    \PassOptionsToPackage{numbers, compress}{natbib}

\usepackage[preprint]{neurips_2021}

\usepackage[utf8]{inputenc} 
\usepackage[T1]{fontenc}    
\usepackage{hyperref}       
\usepackage{url}            
\usepackage{booktabs}       
\usepackage{amsfonts}       
\usepackage{nicefrac}       
\usepackage{microtype}      
\usepackage{multirow}
\usepackage{graphicx}
\usepackage{epsfig}
\usepackage{amsmath}
\usepackage{amssymb}
\usepackage{subfigure}
\usepackage{marvosym}
\usepackage{color}
\usepackage{xcolor}         
\usepackage{threeparttable}
\usepackage{algorithm}
\usepackage{algpseudocode}
\usepackage{setspace}
\usepackage[normalem]{ulem}
\usepackage{amsmath,amssymb,amsfonts}
\usepackage{algorithm}
\usepackage{algorithmicx}
\title{Your time series is worth a binary image:\\ machine vision assisted deep framework for time series forecasting}

\author{%
  Luoxiao Yang, Xinqi Fan, Zijun Zhang (\Letter) \\
  School of Data Science, City University of Hong Kong, Hong Kong SAR\\
  {\texttt{\{luoxiyang2-c,xinqi.fan\}@my.cityu.edu.hk, \{zijzhang\}@cityu.edu.hk}}
}

\begin{document}

\maketitle

\begin{abstract}
Time series forecasting (TSF) has been a challenging research area, and
various models have been developed to address this task. However, almost
all these models are trained with numerical time series data, which is
not as effectively processed by the neural system as visual information.
To address this challenge, this paper proposes a novel machine vision
assisted deep time series analysis (MV-DTSA) framework. The MV-DTSA
framework operates by analyzing time series data in a novel binary
machine vision time series metric space, which includes a mapping and an
inverse mapping function from the numerical time series space to the
binary machine vision space, and a deep machine vision model designed to
address the TSF task in the binary space. A comprehensive computational
analysis demonstrates that the proposed MV-DTSA framework outperforms
state-of-the-art deep TSF models, without requiring sophisticated data
decomposition or model customization. The code for our framework is
accessible at
\url{https://github.com/IkeYang/machine-vision-assisted-deep-time-series-analysis-MV-DTSA-}.
\end{abstract}

\section{Introduction}

Time series refers to a set of data points collected in time order. It
is widespread across various domains such as energy, traffic, health
care, economics, etc. Time series forecasting (TSF) is a long-standing
challenge in the field of time series analysis, aimed at capturing
trends and predicting future values based on historical patterns and
trends.

In literature, TSF models can be divided into two categories:
statistical models and machine learning models. The statistical models
include traditional time series models such as Autoregressive (AR),
Autoregressive Moving Average (ARMA) and Autoregressive Integrated
Moving Average (ARIMA) \cite{hamilton2020time}, while the shallow machine learning
models include models such as neural networks, support vector machines,
decision trees, and ensemble methods \cite{han2019review},\cite{godahewa2021ensembles}. In recent years, deep
learning models, including Recurrent Neural Networks (RNNs) and
Convolutional Neural Networks (CNNs), have gained prominence in TSF due
to their superior performance in modeling complex systems \cite{lecun2015deep}.
Additionally, Transformer architecture \cite{vaswani2017attention}, which benefits from the
attention mechanism and can effectively capture long-term dependencies,
has also been widely adopted in TSF, including Informer \cite{zhou2021informer},
Autoformer \cite{wu2021autoformer}, FEDformer \cite{zhou2022fedformer} and PatchTST \cite{patchTST}.

The existing time series forecasting (TSF) models, whether statistical
or machine learning, have one character in common: they are directly
trained on numerical time series data. However, it is proven that the
human brain processes visual information more efficiently than numerical
data \cite{pettersson1993visual}. Research has demonstrated that the visual cortex is
capable of quickly identifying patterns, shapes, and colors, making
images and videos more quickly processed than text \cite{dondis1974primer}. Given the
deep neural network\textquotesingle s connections to neuroscience
\cite{hassabis2017neuroscience}, it is intriguing to develop a deep architecture for TSF that
leverages the visual patterns of historical time series instead of
relying on numerical values as input to the model.

\begin{figure}[htbp]
    \centering
    \subfigure[Numerical time series data in space \textit{S}]{
        \includegraphics[width=0.5\textwidth]{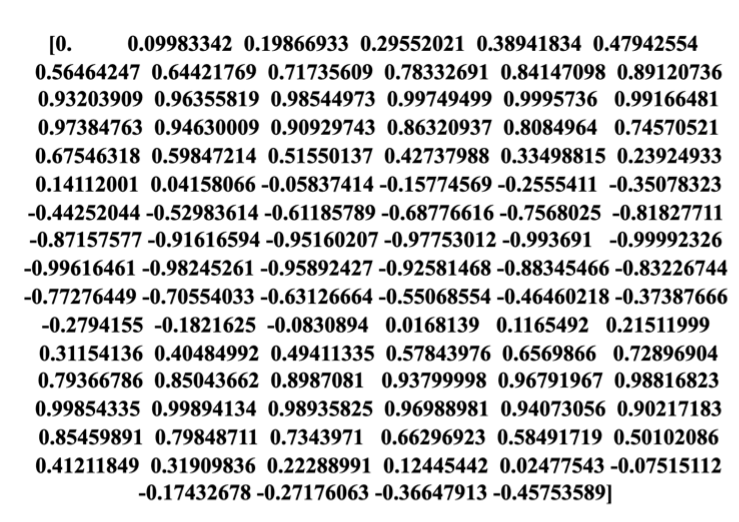}
    }
    \subfigure[Binary machine vision time series data in space \textit{V}]{
	\includegraphics[width=0.4\textwidth]{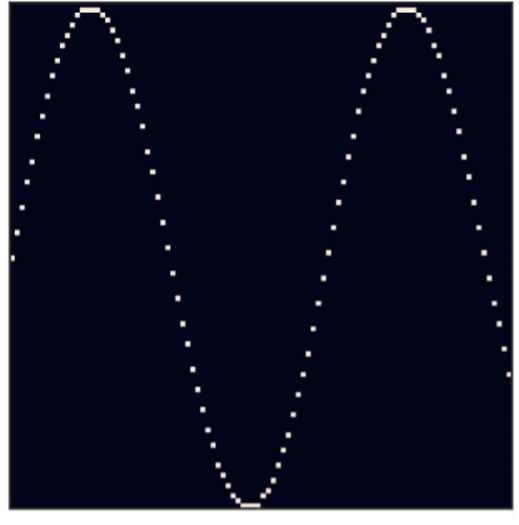}
    }
    \caption{An illustration of numerical time series data and binary machine vision time series data}
    \label{f1}
\end{figure}
In this research, we propose a novel framework for time series analysis,
called machine vision-assisted deep time series analysis (MV-DTSA). The
proposed framework analyzes time series in a binary machine vision time
series metric space, rather than the real space as shown in Fig. \ref{f1}.
After mapping the original one-dimensional numerical time series shown
in Fig. \ref{f1} (a) to a two-dimensional binary tensor shown in Fig. \ref{f1} (b),
only the relative tendency relationship is preserved. The MV-DTSA
framework includes three key steps: (1) defining a binary machine vision
time series metric space \emph{V} and the corresponding mapping and
inverse mapping functions from the numerical time series space \emph{S}
to V; (2) designing deep machine vision models to address the time
series analysis task in \emph{V}; and (3) optionally inverse mapping the
model results back to \emph{S}. A comprehensive computational study has
been conducted to demonstrate the effectiveness of the proposed MV-DTSA
framework applied to TSF task by benchmarking against SOTA deep TSF
models. The results show that the proposed machine vision assist deep
time series forecasting (MV-DTSF) outperforms state-of-the-art deep TSF
models without requiring sophisticated time series decomposition methods
or customized model design, even when using commonly adopted deep
structures.

Main contributions of this work are summarized as follows:
\begin{itemize}
\item[\textbf{1)}]	To leverages the visual patterns, a disruptive framework for time series analyze, the MV-DTSA, is presented for the first time.
\item[\textbf{2)}] A novel MV-DTSF model is designed based on the MV-DTSA.
\item[\textbf{3)}] The proposed MV-DTSF can easily achieve superior prediction performance without sophisticated data decomposition and model customization, which suggests that the proposed framework has significant untapped potential in the field of time series analysis.
\end{itemize}

\vspace{-5pt}
\section{Related Work}
\vspace{-5pt}
\subsection{Models for Time Series Forecasting}
Historically, statistical models such as AR, MA, ARIMA \cite{hamilton2020time}, etc.
have dominated TSF. These models are designed based on rigorous
mathematical derivation and still play an important role today. In
recent years, machine learning techniques, including shallow models such
as LR, SVR, SNN, tree-based methods \cite{han2019review}, and deep models such as
CNN\cite{han2019review}, RNN \cite{han2019review} and its variants \cite{bai2018empirical}, have gained traction
due to their ability to model intricate relationships and patterns in
the data. These techniques have been reported to achieve better
performance than traditional statistical models and are widely used in
many applications. Inspired by the success of the Transformer applied in
NLP \cite{vaswani2017attention} and CV \cite{dosovitskiy2020image}, many attempts have been made in the
literature to apply Transformer to time series forecasting, including
Informer \cite{zhou2021informer}, Autoformer \cite{wu2021autoformer}, FEDformer \cite{zhou2022fedformer}, and patchTST
\cite{patchTST}. These models incorporate various techniques such as
self-attention mechanisms, decomposition architecture, auto-correlation
mechanisms, Fourier transform, wavelet transform, and subseries-level
patches with a channel independent training strategy to improve
performance.

In conclusion, the methods discussed above share three key features.
Firstly, they almost exclusively employ numerical time series values as
inputs. Secondly, sophisticated data preprocessing and decomposition
techniques are frequently adopted to improve the prediction performance.
Lastly, these models, particularly the deep models, have placed great
emphasis on designing the models by carefully crafting the architectures
and loss functions to enhance the accuracy and performance of the TSF
tasks.

In our research, we introduce a disruptive approach by conducting time
series analysis in a binary machine vision time series metric space
instead of directly using numerical time series data. The results
demonstrate that our proposed MV-DTSF model achieves SOTA performance
without relying on sophisticated data decomposition techniques, or
significant emphasis on the design of the models.

\section{Method}
\begin{figure}[!htb]
	\centering{\includegraphics[width=0.8\textwidth]{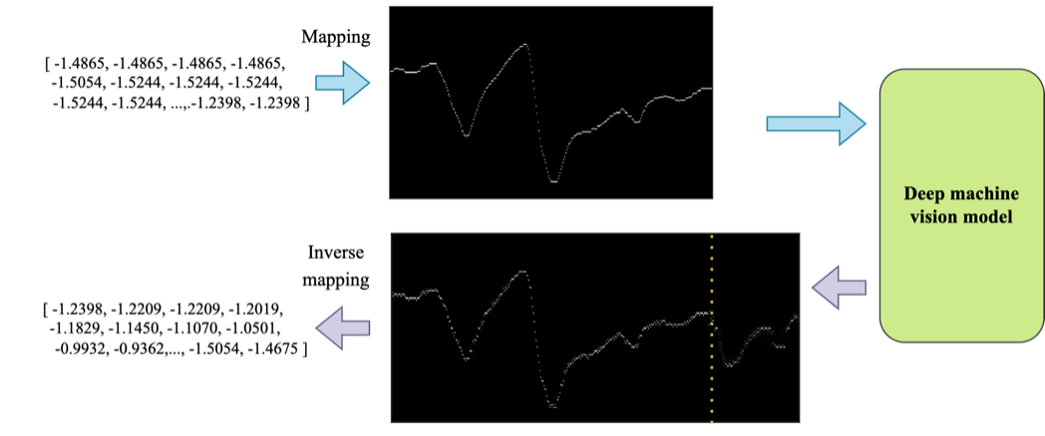}}
	\caption{The overall framework of the proposed MV-DTSA.}
	\label{f2}
\end{figure}
Fig. \ref{f2} depicts the overall framework of the proposed MV-DTSA, which
comprises three essential steps. Firstly, the numerical time series data
from the real space \emph{S} are mapped to the proposed binary machine
vision time series metric space \emph{V}. Secondly, deep machine vision
models are selected based on the requirements of the specific task and
trained using the mapped training pairs on \emph{V}. Lastly, if
necessary, the inverse mapping is performed to translate the output of
the deep model back into the real space for further analysis.

Next, we will delve into the details of the proposed binary machine
vision time series metric space and the MV-DTSF model.

The time series forecasting problem is to predict the most probable length-$O$ series in the future given the past length-$I$ series, denoting as \textit{input-$I$-predict-$O$}. The \textit{long-term forecasting} setting is to predict the long-term future, i.e. larger $O$.
As aforementioned, we have highlighted the difficulties of long-term series forecasting: handling intricate temporal patterns and breaking the bottleneck of computation efficiency and information utilization. To tackle these two challenges, we introduce the decomposition as a builtin block to the deep forecasting model and propose \textit{Autoformer} as a decomposition architecture. Besides, we design the \textit{Auto-Correlation} mechanism to discover the period-based dependencies and aggregate similar sub-series from underlying periods.

\subsection{Binary machine vision time series metric space}

\begin{figure}[!htb]
	\centering{\includegraphics[width=0.8\textwidth]{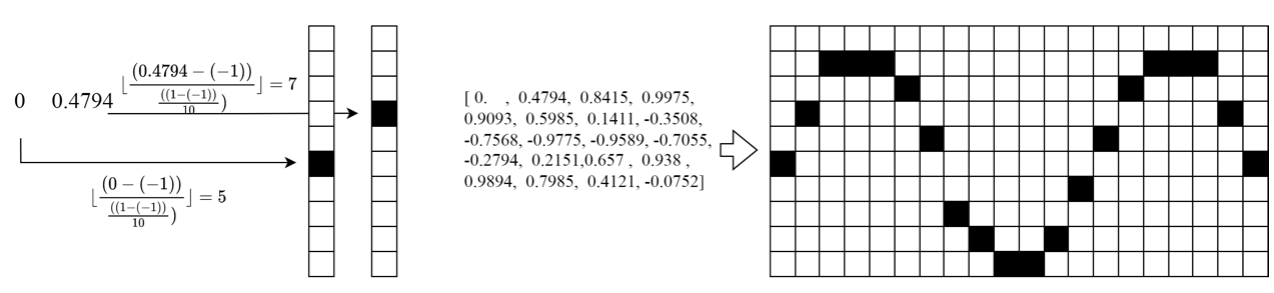}}
	\caption{An illustration of the mapping process from the real space to the binary machine vision time series metric space.}
	\label{f3}
\end{figure}

Firstly, we give the definition of the proposed binary machine vision
time series metric space.

\textbf{Definition 1}: \textbf{Binary machine vision time series metric
space.} The Binary machine vision time series metric space is defined as
a group \((V,\ d)\), where \(V\) is a set of elements as defined in \eqref{eq1}
and \emph{d} is the variant of the earth mover's distance as defined in
\eqref{eq2}.

\begin{equation}
V = \left\{ v \in R^{c \times h \times t} \middle| {\ v}_{i,j,k} \in \left\{ 0,1 \right\},i \in \lbrack c\rbrack,j \in \lbrack h\rbrack,k \in \lbrack t\rbrack,\sum_{j = 1}^{h}{\ v}_{i,j,k} = 1 \right\}\ \\
\label{eq1}
\end{equation}

\begin{equation}
d\left( v_{1},v_{2} \right) = \int_{i = 1}^{c}{\int_{k = 1}^{t}{\inf_{\gamma \in \prod_{}^{}\left( \mathbf{v}_{\mathbf{1}}^{\mathbf{i,}\mathbf{1:h,k}}\mathbf{,}\mathbf{v}_{\mathbf{2}}^{\mathbf{i,}\mathbf{1:h,k}} \right)}{\mathbb{E}_{x,y\sim\gamma}\left\| x - y \right\|_{1}}dkdi}}\ ) \\
\label{eq2}
\end{equation}

\textsubscript{­}where \emph{c} denotes the number of variates, \emph{t}
denotes the length of time series, and \emph{h} denotes the resolution.

In addition, we introduce the mapping function \(f(*)\) from the real
space of numerical time series
\(S = \left\{ s \in R^{c \times t} \middle| s_{i,k} \in R \right\}\ \)to
\(\textit{V}\) and the corresponding inverse mapping function
\(f^{- 1}(*)\) as expressed in \eqref{eq3.1} and \eqref{eq3.2} respectively.

\begin{equation}
\begin{split}
   \boldsymbol{v}_{i, \mathbf{1 : h}, \boldsymbol{k}}=\boldsymbol{f}\left(s_{i, k}\right)&=\left\langle f_1\left(s_{i, k}\right), f_2\left(s_{i, k}\right), \ldots f_h\left(s_{i, k}\right)\right\rangle    \\
\quad f_j\left(s_{i, k}\right) &= \begin{cases}1, & s_{i, k} \geq M S, j=h \\
1, & s_{i, k} \leq M S, j=1 \\
1, & j=\left\lfloor\begin{array}{ll}
\frac{s_{i, k}+M S}{2 M S}
\end{array}\right\rfloor, j \in[h] \\
0, & \text { else }\end{cases} 
\label{eq3.1}
\end{split}
\end{equation}

\begin{equation}
s_{i,k} = f^{- 1}(\mathbf{v}_{\mathbf{i}\mathbf{,1:h,}\mathbf{k}}) = \sum_{j = 1}^{h}{((j - 0.5) \times \frac{2MS}{h} - MS) \times v_{i,j,k}}\ \\
\label{eq3.2}
\end{equation}

where \(MS > 0\) denotes the maximum scale hyperparameter. Typically,
before mapping, numerical time series in real space \emph{S} undergo
zero-score normalization. After mapping to the binary machine vision
time series metric space \emph{V}, only the spatial-temporal tendency of
the original time series is preserved. An illustration of the mapping
function from \emph{S} to \emph{V} is provided in Fig. 3, where
\emph{MS}=1 and \emph{h}=10. It is observable that, after mapping, the
original time series are converted to a one-channel binary image.
Meanwhile, it is evident that there exists system measurement error
(SME) in the mapping and the inverse mapping process, which is
formalized by \textbf{Theorem 1}.

\vspace{-5pt}
\paragraph{Assumption 1}: After zero-score normalization,
\(S\sim N(\mathbf{0,I})\).

\vspace{-5pt}
\paragraph{Theorem 1}: Let \(\widehat{s} \in S\), the SME is defined as
\(\left\| f^{- 1}\left( \mathbf{f}\left( \widehat{s} \right) \right) - \widehat{s} \right\|_{1}\ \)then
the expectation of SME can be bounded as:

\begin{equation}
\begin{split}
    &\mathbb{E}\left\| f^{- 1}\left( \mathbf{f}\left( \widehat{s} \right) \right) - \widehat{s} \right\|_{1}  \leq g(h,MS) \\
&= ct\left( MS\left( \frac{1}{h}(\Phi(MS) - \Phi( - MS)) - 2 + 2\Phi(MS) \right) + \sqrt{\frac{2}{\pi}}e^{\frac{- MS^{2}}{2}} \right)
\label{eq4}
\end{split}
\end{equation}

where \(\Phi\) denotes the cumulative density function of
\(N(\mathbf{0,I})\) and \(P(*)\) denotes the probability density
function of \(N(0,1)\).

Denote
\(MS\left( \frac{1}{h}(\Phi(MS) - \Phi( - MS)) - 2 + 2\Phi(MS) \right) + \sqrt{\frac{2}{\pi}}e^{\frac{- MS^{2}}{2}}\)
in \eqref{eq4} as the upper bound of SME. It is clear that when the \emph{MS} is fixed, the upper bound of SME will decrease with the increase of the
\emph{h}. Then we have \textbf{Proposition 1} to guarantee the
convergence.

\vspace{-5pt}
\paragraph{Proposition 1}: When \(h \longrightarrow + \infty\),
\(\forall\varepsilon > 0,\ \exists\delta,\ \forall MS \in R^{+}\)and\(\ \forall MS \geq \delta\),
\eqref{eqP1} holds.

\begin{equation}
\left| MS\left( \frac{1}{h}\left( \Phi(MS) - \Phi( - MS) \right) - 2 + 2\Phi(MS) \right) + \sqrt{\frac{2}{\pi}}e^{\frac{- MS^{2}}{2}} \right| \leq \varepsilon\ \\
\label{eqP1}
\end{equation}

\textbf{Proposition 1} guarantees the convergence of the upper bound of
the SME. Specifically, if an adequate amount of GPU memory is available
to specify a sufficiently large h, then SME will converge to 0 given a
large enough value of MS. In practice, however, a large value of h
results in a larger tensor size within the binary machine vision time
series metric space, which can lead to an increase in computational
load. Consequently, the selection of h should be carefully chosen to
ensure computational feasibility. Then we have \textbf{Proposition 2}.

\vspace{-5pt}
\paragraph{Proposition 2}: Given h, there always exists a best \(MS^{*}\)
satisfied (6) to the minimize upper bound of SME.

\begin{equation}
\frac{1}{h}\left( \Phi\left( MS^{*} \right) - \Phi\left( - MS^{*} \right) \right) - 2 + 2\Phi\left( MS^{*} \right) + \frac{MS^{*}}{h}\frac{\sqrt{2}}{\sqrt{\pi}}e^{- \frac{{MS^{*}}^{2}}{2}} = 0\
\label{eqP2}
\end{equation}

Finding an analytic solution for equation \eqref{eqP2} is challenging. As a result, numerical solutions are typically employed in practice. Table \ref{t1} presents the numerical results of the optimal MS obtained for various h values.

\begin{table}
\centering
\caption{The numerical results of the optimal  and the corresponding upper bound of SME given different \emph{h}.}
\label{t1}
\begin{tabular}{ccc} 
\toprule
\emph{h}                            & Best \emph{MS}   & Upper bound of SME  \\ \hline
50 & 2.29 & 0.052 \\
100 & 2.55 & 0.028 \\
200 & 2.79 & 0.015 \\
400 & 3.02 & 0.008 \\
800 & 3.22 & 0.004 \\
\bottomrule
\end{tabular}
\end{table}

\subsection{Machine vision assist deep time series forecasting}

\subsubsection{The overall architecture of the proposed MV-DTSF.}

\begin{figure}[!htb]
	\centering{\includegraphics[width=0.8\textwidth]{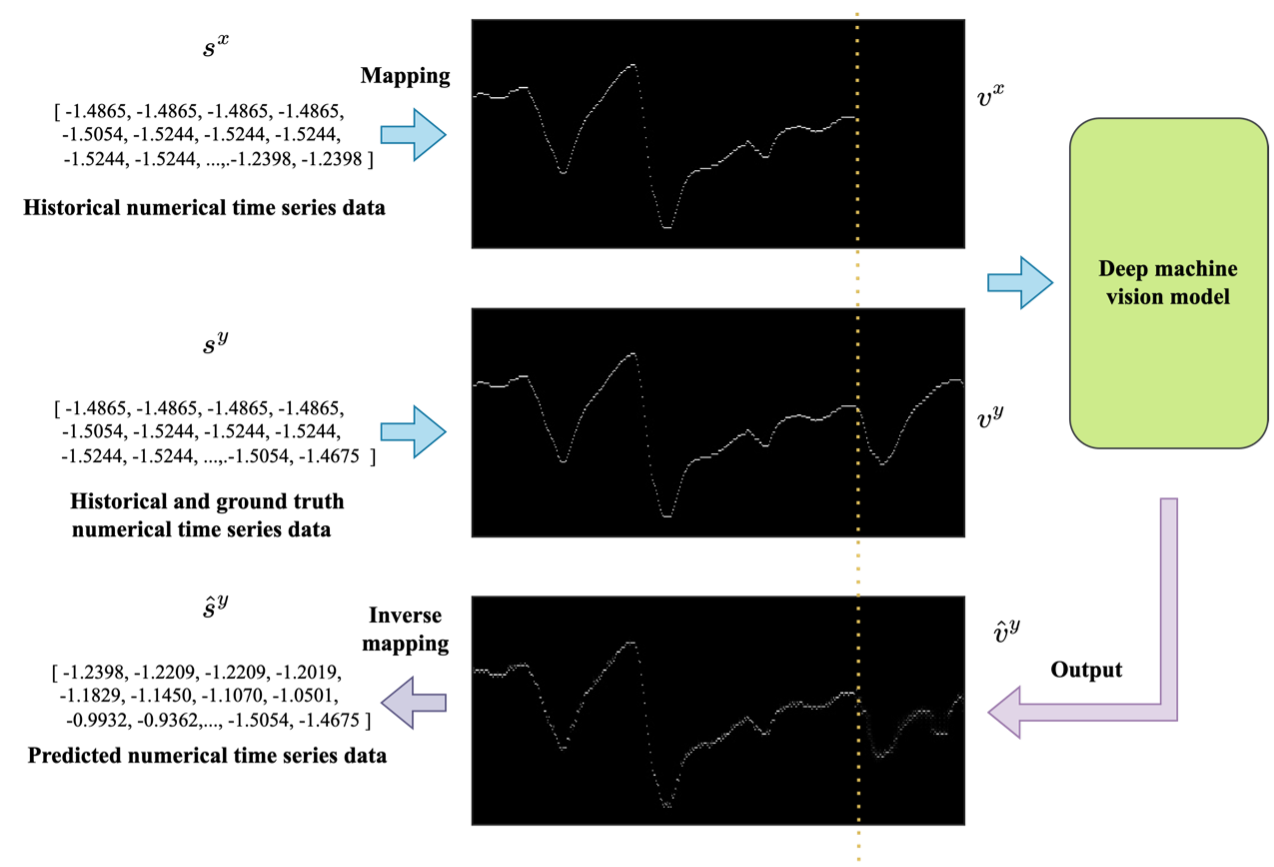}}
	\caption{The architecture of the proposed MV-DTSF.}
	\label{f4}
\end{figure}

To evaluate the effectiveness of the proposed MV-DTSA framework, we
apply it to TSF task. The overall architecture of the proposed MV-DTSF
is depicted in Fig. \ref{f4}. Initially, the historical numerical time series
data, \(\mathbf{s}^{\mathbf{x}}\), and the ground truth numerical time
series data, \(\mathbf{s}^{\mathbf{y}}\), are mapped into the binary
machine vision time series metric space, \emph{V}, as outlined in
equation \eqref{eq8}.

\begin{equation}
\begin{matrix}
\mathbf{v}^{\mathbf{x}}=\mathbf{f}\left( \mathbf{s}^{\mathbf{x}} \right) & \mathbf{v}^{\mathbf{y}}=\mathbf{f}\left( \mathbf{s}^{\mathbf{y}} \right)
\end{matrix}
\label{eq8}
\end{equation}

Next, an end-to-end deep machine vision model \emph{D} are adopted and
trained with \(\mathbf{v}^{\mathbf{x}}\) and \(\mathbf{v}^{\mathbf{y}}\)
to output the predicted results \({\widehat{\mathbf{v}}}^{\mathbf{y}}\):

\begin{equation}
{\widehat{\mathbf{v}}}^{\mathbf{y}} = D\left( \mathbf{v}^{\mathbf{x}} \right)
\label{eq9}
\end{equation}

The objective function of MV-DTSF is trying to minimize the distance
better \({\widehat{\mathbf{v}}}^{\mathbf{y}}\) and
\(\mathbf{v}^{\mathbf{y}}\) as expressed in \eqref{eq10}:

\begin{equation}
\theta^{*} = \underset{\theta}{\mathrm{argmin}}\ d\left( {\widehat{\mathbf{v}}}^{\mathbf{y}},\mathbf{v}^{\mathbf{y}} \right)
\label{eq10}
\end{equation}

where \(\theta\) denotes parameters of the deep machine vision model
\emph{D} and \emph{d} denotes the distance function defined in \eqref{eq2}.

After training, the testing data is fed into the developed deep machine
vision model and the corresponding output
\({\widehat{\mathbf{v}}}^{\mathbf{y}}\) is inverse mapped to
\({\widehat{\mathbf{s}}}^{\mathbf{y}}\) for further analysis:

\begin{equation}
{\widehat{\mathbf{s}}}^{\mathbf{y}}=f^{-1}\left({\widehat{\mathbf{v}}}^{\mathbf{y}}\right)
\label{eq11}
\end{equation}
Pseudocode of the training and testing process are offered in
\ref{alg:train}.

\begin{algorithm}[H]    
    \caption{Training \& testing process of the proposed
MV-DTSF}\label{alg:train}
    \begin{algorithmic}[1]
        \Require{Historical numerical time series data
\(\mathbf{s}^{\mathbf{x}}\), ground truth numerical time series data
\(\mathbf{s}^{\mathbf{y}}\), hyperparameter \emph{h.}}
        
        \Statex \textbf{Training}
        \State Initialize weights of the deep machine vision model D.
        \State Find the best MS by solving \eqref{eqP2}
        \State $\mathbf{v}^{\mathbf{x}}\mathbf{=}\mathbf{f}\left( \mathbf{s}^{\mathbf{x}} \right)$ \quad
        $ \mathbf{v}^{\mathbf{y}}\mathbf{=}\mathbf{f}\left( \mathbf{s}^{\mathbf{y}} \right)$ 
        \For{$i=1$ to $n$}
        
        \State ${\widehat{\mathbf{v}}}^{\mathbf{y}} = D\left( \mathbf{v}^{\mathbf{x}} \right)$
        \State calculate the loss function based on (2) and update \emph{D}
        \EndFor
        \Statex \textbf{Testing}
        \State $\mathbf{v}^{\mathbf{x}}\mathbf{=}\mathbf{f}\left( \mathbf{s}^{\mathbf{x}} \right)$      
        \State ${\widehat{\mathbf{v}}}^{\mathbf{y}} = D\left( \mathbf{v}^{\mathbf{x}} \right)$
        \State ${\widehat{\mathbf{s}}}^{\mathbf{y}}\mathbf{=}f^{- 1}\left( {\widehat{\mathbf{v}}}^{\mathbf{y}} \right)$
        \State \textbf{Return} $\widehat{\mathbf{s}}^{\mathbf{y}}$
        
    \end{algorithmic}
\end{algorithm}

\subsubsection{Training details.}

\vspace{-5pt}
\paragraph{Deep machine vision models}

To align with the proposed MV-DTSF, we consider the deep machine vision
models utilized in vision segmentation task that can take an image as
input and generate an output image of the same size. For this purpose,
two classic image segmentation models, the U-net \cite{falk2019unet} and the
DeepLabV2 \cite{chen2017deeplab}, are selected as the deep machine vision models.

\vspace{-5pt}
\paragraph{In-sequence data normalization}

To meet the requirements of \textbf{Assumption 1}, we perform an
in-sequence data normalization rather than normalizing the entire
sequence, as described in \eqref{eqInseqNor}:

\begin{equation}
\begin{matrix}
\mathbf{s}^{\mathbf{x}} = \frac{\mathbf{s}^{\mathbf{x}} - mean\left( \mathbf{s}^{\mathbf{x}} \right)}{std\left( \mathbf{s}^{\mathbf{x}} \right)} & \mathbf{s}^{\mathbf{y}} = \frac{\mathbf{s}^{\mathbf{y}} - mean\left( \mathbf{s}^{\mathbf{x}} \right)}{std\left( \mathbf{s}^{\mathbf{x}} \right)} \\
\end{matrix}
\label{eqInseqNor}
\end{equation}

where \(mean(*)\) and \(std(*)\) denote the arithmetic mean function and
the standard deviation function.

Based on \eqref{eqInseqNor}, it is noteworthy that the input numerical time series data \(\mathbf{s}^{\mathbf{x}}\) and \(\mathbf{s}^{\mathbf{y}}\) are
normalized based on the arithmetic mean and standard deviation of
\(\mathbf{s}^{\mathbf{x}}\) rather than the whole sequence.

\vspace{-5pt}
\paragraph{Channel independent}

In the \cite{patchTST}, the author found that dividing multivariate time series
data into separate channels and feeding them independently to the same
model backbone can improve prediction accuracy. In this paper, we adopt
the same channel independent technique used in \cite{patchTST} to ensure a fair comparison in our proposed MV-DTSF.

\section{Computational experiments}

\subsection{Datasets}
Six popular open datasets, Electricity, Exchange rate, Traffic, Weather, ILI and ETTm2 are utilized to evaluate the effectiveness of the proposed MV-DTSF. Details of six datasets are reported in Table \ref{t2}.

\begin{table}
\centering
\caption{Descriptions of datasets in this study}
\label{t2}
\begin{tabular}{ccccccc} 
\toprule
Dataset  & Electricity & Exchange rate &  Traffic &Weather &ILI &ETTm2 \\
\hline 
Features & 321 & 8 & 862 & 21 & 7 & 7 \\
Timesteps & 26304 & 7588 & 17544 & 52696 & 966 & 69680 \\
\bottomrule
\end{tabular}
\end{table}

\subsection{Benchmarks and model setup}
To demonstrate the superiority of the proposed MV-DTSF method, five
recent SOTA time series forecasting models are selected as the
benchmarks in this paper, including Informer \cite{zhou2021informer}, Autoformer \cite{wu2021autoformer}, FEDformer \cite{zhou2022fedformer}, patchTST \cite{patchTST} and DLinear \cite{dlinear}. The
prediction length T is set to \{24, 36, 48, 60\} for the ILI dataset and
\{96, 192, 336, 720\} for all other datasets, as reported in \cite{wu2021autoformer}. To
ensure a fair comparison, we categorize the considered benchmarks into
three groups based on the original length of the look-back window and
the use of the channel independent technique. The grouping results are
presented in Table \ref{t3}. Furthermore, four variants of the MV-DTSF method
are proposed in Table \ref{t3}. For example, MV-DTSF-Unet-96 implies the use
of U-net as the backbone of the deep machine vision model with a
look-back window of 96, and MV-DTSF-DeeplabV2-336-CI implies the use of
DeeplabV2 as the backbone of the deep machine vision model with a
look-back window of 336 and the adoption of the channel independent
technology. To alleviate the computational burden, we only consider
DeeplabV2 as the backbone in group 2 and 3. Finally, we set \emph{h} of
all MV-DTSF models to 200 due to computational resource limitations.

\begin{table}
\centering
\caption{Descriptions of datasets in this study}
\label{t3}
\begin{threeparttable}
\begin{tabular}{cccc} 
\toprule
Methods  & Group & Look-back window &  Channel independent \\
\hline 
FEDformer & 1 & 96 (36)* & \textbf{×} \\
Autoformer & 1 & 96 (36)* & \textbf{×} \\
Informer & 1 & 96 (36)* & \textbf{×} \\
PatchTST/42 & 2 & 336 (104)* & $\surd$ \\
DLinear & 2 & 336 (104)* & $\surd$ \\
PatchTST/64 & 3 & 512 (104)* & $\surd$ \\
MV-DTSF-Unet-96 & 1 & 96 (36)* & \textbf{×} \\
MV-DTSF-DeeplabV2-96 & 1 & 96 (36)* & \textbf{×} \\
MV-DTSF-DeeplabV2-336-CI & 2 & 336 (104)* & $\surd$ \\
MV-DTSF-DeeplabV2-512-CI & 3 & 512 & $\surd$ \\
\bottomrule
\end{tabular}
\begin{tablenotes} 
		\item *Look-back window for ILI dataset. 
\end{tablenotes} 
\end{threeparttable}
\end{table}

\subsection{Computational results}

\begin{figure}[!htb]
	\centering{\includegraphics[width=0.5\textwidth]{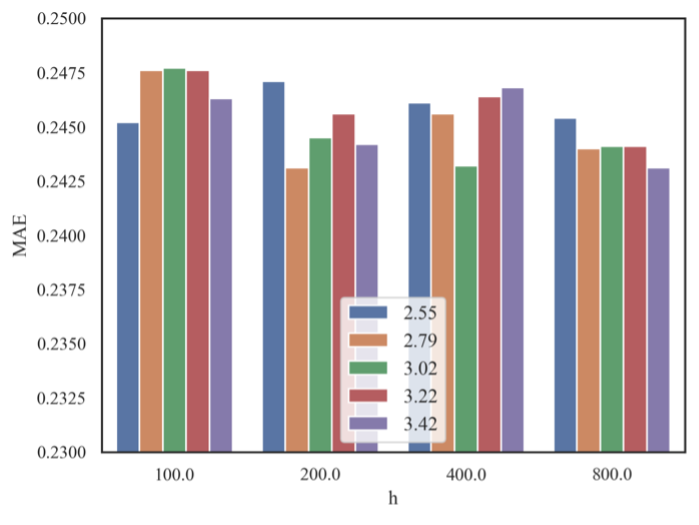}}
	\caption{Forecasting performance (MAE, prediction length=96) with varying \emph{MS} on dataset ETTm2 of model MV-DTSF-DeeplabV2-336-CI.}
	\label{f5}
\end{figure}

\begin{figure}[!htb]
	\centering{\includegraphics[width=0.5\textwidth]{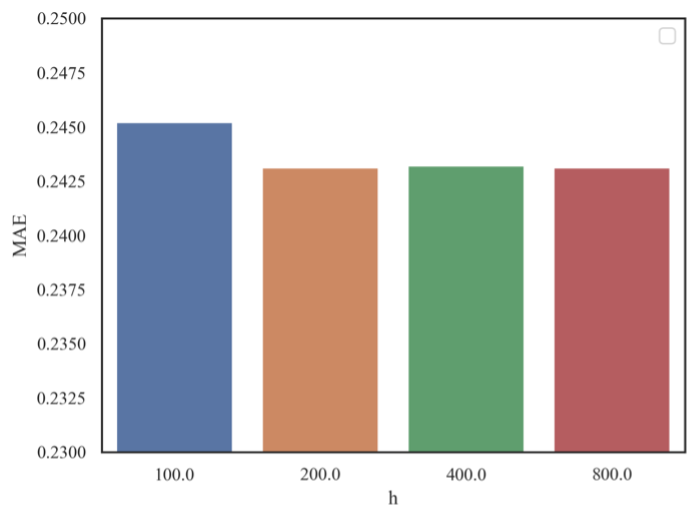}}
	\caption{Forecasting performance (MAE, prediction length=96) with varying \emph{h} on dataset ETTm2 of model MV-DTSF-DeeplabV2-336-CI}
	\label{f6}
\end{figure}

Fig. \ref{f5} displays the prediction performance of the
MV-DTSF-DeeplabV2-336-CI model with different values of h on dataset
ETTm2 for varying MS. It is observable that the best MS results align
closely with the theoretical results reported in Table \ref{t3}. Fig. \ref{f6}
illustrates the prediction performance of the MV-DTSF-DeeplabV2-336-CI
with different h on dataset ETTm2. It is apparent that increasing the
value of h leads to a decrease in prediction error, which supports
\textbf{Proposition 1} and further demonstrates the effectiveness of the
proposed MV-DTSA framework. Notably, based on the observations made from
Figs \ref{f5}-\ref{f6}, our results suggest that for large values of h, such as 800, the error between the theoretical analysis and the observed forecasting
performance appears to be enlarged. This may be attributed to the
sparsity of input data, which causes the deep model to fail in fitting
the data effectively.

\begin{table}
\centering
\caption{The forecasting performance of considered methods of group 1. The best results are in \textbf{bold} and the second best are \uline{underlined}.}
\label{t4}
\renewcommand{\arraystretch}{1.5}
\scalebox{0.8}{
\begin{threeparttable}
\begin{tabular}{l|l|ll|ll|ll|ll|ll|ll} 
\hline
\multicolumn{2}{l|}{Method}        & \multicolumn{2}{l|}{\begin{tabular}[c]{@{}l@{}}MV-DTSF\\-Unet-96\end{tabular}} & \multicolumn{2}{l|}{\begin{tabular}[c]{@{}l@{}}MV-DTSF\\-DeeplabV2-96\end{tabular}} & \multicolumn{2}{l|}{FEDformer-f*} & \multicolumn{2}{l|}{FEDformer-w*} & \multicolumn{2}{l|}{Autoformer*} & \multicolumn{2}{l}{Informer*}  \\ 
\hline
\multicolumn{2}{l|}{Metric}        & MSE            & MAE                                                           & MSE            & \multicolumn{1}{l}{MAE}                                            & MSE           & MAE               & MSE            & MAE              & MSE   & MAE                      & MSE   & MAE                    \\ 
\hline
\multirow{4}{*}{\rotatebox{90}{Electricity} }& 96  & \multicolumn{2}{l|}{\multirow{4}{*}{NAN}}                                      & \textbf{0.167} & \textbf{0.265}                                                     & 0.193         & 0.308             & \uline{0.183}  & \uline{0.297}    & 0.201 & 0.317                    & 0.274 & 0.368                  \\
                             & 192 & \multicolumn{2}{l|}{}                                                          & \textbf{0.180} & \textbf{0.276}                                                     & 0.201         & 0.315             & \uline{0.195}  & \uline{0.308}    & 0.222 & 0.334                    & 0.296 & 0.386                  \\
                             & 336 & \multicolumn{2}{l|}{}                                                          & \textbf{0.190} & \textbf{0.286}                                                     & 0.214         & 0.329             & \uline{0.212}  & \uline{0.313}    & 0.231 & 0.338                    & 0.300 & 0.364                  \\
                             & 720 & \multicolumn{2}{l|}{}                                                          & \textbf{0.220} & \textbf{0.310}                                                     & 0.246         & 0.355             & \uline{0.231}  & \uline{0.343}    & 0.254 & 0.361                    & 0.373 & 0.439                  \\ 
\hline
\multirow{4}{*}{\rotatebox{90}{Exchange} }   & 96  & \textbf{0.108} & \textbf{0.229}                                                & \uline{0.112}  & \uline{0.237}                                                      & 0.148         & 0.278             & 0.139          & 0.276            & 0.197 & 0.323                    & 0.847 & 0.752                  \\
                             & 192 & \uline{0.207}  & \textbf{0.319}                                                & \textbf{0.198} & \uline{0.321}                                                      & 0.271         & 0.380             & 0.256          & 0.369            & 0.300 & 0.369                    & 1.204 & 0.895                  \\
                             & 336 & \textbf{0.304} & \textbf{0.407}                                                & \uline{0.358}  & \uline{0.434}                                                      & 0.460         & 0.500             & 0.426          & 0.464            & 0.509 & 0.524                    & 1.672 & 1.036                  \\
                             & 720 & \uline{0.924}  & \uline{0.729}                                                 & \textbf{0.913} & \textbf{0.723}                                                     & 1.195         & 0.841             & 1.090          & 0.800            & 1.145 & 0.941                    & 2.478 & 1.310                  \\ 
\hline
\multirow{4}{*}{\rotatebox{90}{Traffic}}     & 96  & \multicolumn{2}{l|}{\multirow{4}{*}{NAN}}                                      & 0.623          & \textbf{0.311}                                                     & \uline{0.587} & 0.366             & \textbf{0.562} & \uline{0.349}    & 0.613 & 0.388                    & 0.719 & 0.391                  \\
                             & 192 & \multicolumn{2}{l|}{}                                                          & 0.637          & \textbf{0.322}                                                     & \uline{0.604} & 0.373             & \textbf{0.562} & \uline{0.346}    & 0.616 & 0.382                    & 0.696 & 0.379                  \\
                             & 336 & \multicolumn{2}{l|}{}                                                          & 0.652          & \textbf{0.327}                                                     & \uline{0.621} & 0.383             & \textbf{0.570} & \uline{0.323}    & 0.622 & 0.337                    & 0.777 & 0.420                  \\
                             & 720 & \multicolumn{2}{l|}{}                                                          & 0.689          & \textbf{0.345}                                                     & \uline{0.626} & 0.382             & \textbf{0.596} & \uline{0.368}    & 0.660 & 0.408                    & 0.864 & 0.472                  \\ 
\hline
\multirow{4}{*}{\rotatebox{90}{Weather} }    & 96  & \uline{0.167}  & \uline{0.205}                                                 & \textbf{0.167} & \textbf{0.205}                                                     & 0.217         & 0.296             & 0.227          & 0.304            & 0.266 & 0.336                    & 0.300 & 0.384                  \\
                             & 192 & \uline{0.238}  & \uline{0.271}                                                 & \textbf{0.217} & \textbf{0.252}                                                     & 0.276         & 0.336             & 0.295          & 0.363            & 0.307 & 0.367                    & 0.598 & 0.544                  \\
                             & 336 & \uline{0.296}  & \uline{0.314}                                                 & \textbf{0.282} & \textbf{0.303}                                                     & 0.339         & 0.380             & 0.381          & 0.416            & 0.359 & 0.395                    & 0.578 & 0.523                  \\
                             & 720 & \uline{0.371}  & \uline{0.362}                                                 & \textbf{0.362} & \textbf{0.352}                                                     & 0.403         & 0.428             & 0.424          & 0.434            & 0.419 & 0.428                    & 1.059 & 0.741                  \\ 
\hline
\multirow{4}{*}{\rotatebox{90}{ILI} }        & 24  & \uline{1.915}  & \uline{0.829}                                                 & \textbf{1.782} & \textbf{0.810}                                                     & 3.228         & 1.260             & 2.203          & 0.963            & 3.483 & 1.287                    & 5.764 & 1.677                  \\
                             & 36  & \textbf{1.993} & \textbf{0.840}                                                & \uline{2.460}  & \uline{0.894}                                                      & 2.679         & 1.080             & 2.272          & 0.976            & 3.103 & 1.148                    & 4.755 & 1.467                  \\
                             & 48  & \textbf{1.800} & \textbf{0.804}                                                & \uline{1.980}  & \uline{0.820}                                                      & 2.622         & 1.078             & 2.209          & 0.987            & 2.669 & 1.085                    & 4.763 & 1.469                  \\
                             & 60  & \textbf{1.406} & \textbf{0.752}                                                & \uline{1.537}  & \uline{0.791}                                                      & 2.857         & 1.157             & 2.545          & 1.061            & 2.770 & 1.125                    & 5.264 & 1.564                  \\ 
\hline
\multirow{4}{*}{\rotatebox{90}{ETTm2}}       & 96  & \uline{0.189}  & \uline{0.262}                                                 & \textbf{0.182} & \textbf{0.258}                                                     & 0.203         & 0.287             & 0.204          & 0.288            & 0.255 & 0.339                    & 0.365 & 0.453                  \\
                             & 192 & \uline{0.260}  & \uline{0.313}                                                 & \textbf{0.247} & \textbf{0.301}                                                     & 0.269         & 0.328             & 0.346          & 0.363            & 0.281 & 0.340                    & 0.533 & 0.563                  \\
                             & 336 & \uline{0.324}  & \uline{0.354}                                                 & \textbf{0.302} & \textbf{0.338}                                                     & 0.325         & 0.366             & 0.359          & 0.387            & 0.339 & 0.372                    & 1.363 & 0.887                  \\
                             & 720 & \uline{0.427}  & \uline{0.412}                                                 & \textbf{0.399} & \textbf{0.396}                                                     & 0.421         & 0.415             & 0.433          & 0.432            & 0.422 & 0.419                    & 3.379 & 1.338                  \\
\hline
\end{tabular}
      \begin{tablenotes} 
		\item * Results are from FEDformer \cite{zhou2022fedformer}. 
     \end{tablenotes} 
\end{threeparttable} 
}
\end{table}

\begin{table}
\centering
\caption{The forecasting performance of considered methods of group 2 and 3. The best results are in \textbf{bold} and the second best is \uline{underlined}.}
\label{t5}
\renewcommand{\arraystretch}{1.5}
\scalebox{0.8}{
\begin{threeparttable}

\begin{tabular}{l|l|ll|ll|ll|ll|ll} 
\hline
\multicolumn{2}{l|}{Method}    & \multicolumn{2}{l|}{\begin{tabular}[c]{@{}l@{}}MV-DTSF\\-DeeplabV2-512-CI\end{tabular}} & \multicolumn{2}{l|}{patchTST/64 *} & \multicolumn{2}{l|}{\begin{tabular}[c]{@{}l@{}}MV-DTSF\\-DeeplabV2-336-CI\end{tabular}} & \multicolumn{2}{l}{patchTST/42*} & \multicolumn{2}{|l}{Dlinear*}  \\ 
\hline
\multicolumn{2}{l|}{Metric}    & MSE           & MAE                                                                     & MSE            & MAE               & MSE           & MAE                                                                     & MSE            & MAE             & MSE   & MAE                   \\ 
\hline
\multirow{4}{*}{\rotatebox{90}{Weather}} & 96  & 0.154         & \uline{0.191}                                                           & \textbf{0.149} & 0.198             & \uline{0.150} & \textbf{0.189}                                                          & 0.152          & 0.199           & 0.176 & 0.237                 \\
                         & 192 & 0.204         & \uline{0.235}                                                           & \textbf{0.194} & 0.241             & 0.199         & \textbf{0.233}                                                          & \uline{0.197}  & 0.243           & 0.220 & 0.282                 \\
                         & 336 & 0.269         & 0.285                                                                   & \textbf{0.245} & \uline{0.282}     & 0.256         & \textbf{0.276}                                                          & \uline{0.249}  & 0.283           & 0.265 & 0.319                 \\
                         & 720 & 0.340         & 0.338                                                                   & \textbf{0.314} & \textbf{0.334}    & 0.342         & 0.339                                                                   & \uline{0.320}  & 0.335           & 0.323 & 0.362                 \\ 
\hline
\multirow{4}{*}{\rotatebox{90}{ILI}  }   & 24  & 1.695         & \uline{0.767}                                                           & \textbf{1.319} & \textbf{0.754}    & \multicolumn{2}{l|}{\multirow{4}{*}{NAN}}                                               & \uline{1.522}  & 0.814           & 2.215 & 1.081                 \\
                         & 36  & 1.675         & \textbf{0.776}                                                          & \uline{1.579}  & 0.870             & \multicolumn{2}{l|}{}                                                                   & \textbf{1.430} & \uline{0.834}   & 1.963 & 0.963                 \\
                         & 48  & \uline{1.650} & \textbf{0.773}                                                          & \textbf{1.553} & \uline{0.815}     & \multicolumn{2}{l|}{}                                                                   & 1.673          & 0.854           & 2.130 & 1.024                 \\
                         & 60  & \uline{1.517} & \textbf{0.759}                                                          & \textbf{1.470} & \uline{0.788}     & \multicolumn{2}{l|}{}                                                                   & 1.529          & 0.862           & 2.368 & 1.096                 \\ 
\hline
\multirow{4}{*}{\rotatebox{90}{ETTm2} }  & 96  & 0.167         & \textbf{0.243}                                                          & \uline{0.166}  & 0.256             & 0.170         & \uline{0.244}                                                           & \textbf{0.165} & 0.255           & 0.167 & 0.260                 \\
                         & 192 & 0.234         & \uline{0.287}                                                           & \uline{0.223}  & 0.296             & 0.227         & \textbf{0.286}                                                          & \textbf{0.220} & 0.292           & 0.224 & 0.303                 \\
                         & 336 & 0.287         & \uline{0.326}                                                           & \textbf{0.274} & 0.329             & 0.281         & \textbf{0.321}                                                          & \uline{0.278}  & 0.329           & 0.281 & 0.342                 \\
                         & 720 & 0.376         & 0.389                                                                   & \textbf{0.362} & \uline{0.385}     & 0.379         & \textbf{0.384}                                                          & \uline{0.367}  & 0.385           & 0.397 & 0.421                 \\
\hline
\end{tabular}

      \begin{tablenotes} 
		\item * Results are from patchTST \cite{patchTST}. 
     \end{tablenotes} 
\end{threeparttable} 
}
\end{table}

Table \ref{t4} and Table \ref{t5} report the forecasting performance of the methods in groups 1, 2, and 3, respectively. However, due to computational
constraints, the forecasting performance of MV-DTSF-Unet-96,
MV-DTSF-DeeplabV2-512-CI, and MV-DTSF-DeeplabV2-336-CI in the
Electricity and Traffic datasets are not included. It is observable that
our proposed MV-DTSF method outperforms all the considered benchmarks in
group 1 and achieves comparable performance in groups 2 and 3. It is
worth noting that the proposed method exhibits excellent performance in
terms of MAE suggesting its robustness in handling anomalies. Examples
of the proposed MV-DTSF\textquotesingle s prediction results are
presented in Appendix \ref{appendix:illstruction}.

\section{Conclusion}
In this paper, we proposed a novel framework for time series analysis,
the MV-DTSA, which analyze time series data in the novel binary machine
vision time series metric space. The proposed framework consisted of
three steps: (1) mapping numerical time series data to the proposed
binary machine vision time series metric space, (2) designing a deep
machine vision model and training it with the mapped data, and (3) if
needed, inverse mapping the results from the binary machine vision time
series metric space to the numerical time series data space.
Experimental results on TSF task demonstrated the effectiveness of the
proposed MV-DTSA framework by benchmarking against SOTA TSF models.

In summary, this paper presented a novel approach for time series data
analysis, i.e., analyzing time series from the machine vision
perspective. The proposed method could be applied to various time series
tasks beyond time series forecasting. In the further, we would like to
further improve the proposed MV-DTSA framework by customizing the deep
machine vision model and refining the definition of the binary machine
vision time series metric space.

\small
\bibliographystyle{plain}
\bibliography{ref}

\appendix
\section{Proof of Theorem and Propositions}\label{appendix:proof}
\vspace{-5pt}
\paragraph{Theorem 1}: Let \(\widehat{s} \in S\), the SME is defined as
\(\left\| f^{- 1}\left( \mathbf{f}\left( \widehat{s} \right) \right) - \widehat{s} \right\|_{1}\ \)then
the expectation of SME can be bounded as:

\begin{equation*}
\begin{split}
    &\mathbb{E}\left\| f^{- 1}\left( \mathbf{f}\left( \widehat{s} \right) \right) - \widehat{s} \right\|_{1}  \leq g(h,MS) \\
&= ct\left( MS\left( \frac{1}{h}(\Phi(MS) - \Phi( - MS)) - 2 + 2\Phi(MS) \right) + \sqrt{\frac{2}{\pi}}e^{\frac{- MS^{2}}{2}} \right)
\end{split}
\end{equation*}

where \(\Phi\) denotes the cumulative density function of
\(N(\mathbf{0,I})\) and \(P(*)\) denotes the probability density
function of \(N(0,1)\).

\emph{Proof}:
\begin{equation}
\mathbb{E}\left\| f^{- 1}\left( \mathbf{f}\left( \widehat{s} \right) \right) - \widehat{s} \right\|_{1} = \mathbb{E}\left( \sum_{i}^{c}{\sum_{k}^{t}\left| f^{- 1}\left( \mathbf{f}\left( {\widehat{s}}_{i,k} \right) \right) - {\widehat{s}}_{i,k} \right|} \right) = ct\int_{- \infty}^{+ \infty}{\left| f^{- 1}\left( \mathbf{f}(s) \right) - s \right|P(s)d(s)}
\label{eqA1}
\end{equation}

Based on \eqref{eq3.1}, if \(s \geq MS\), we have
\(\left| f^{- 1}\left( \mathbf{f}(s) \right) - s \right| = s - MS\) and
if \(s \leq - MS\), we have
\(\left| f^{- 1}\left( \mathbf{f}(s) \right) - s \right| = - MS - s\).
Thus, if \(|s| \geq MS\), we have
\(\left| f^{- 1}\left( \mathbf{f}(s) \right) - s \right| = |s| - MS\).
Considering Assumption 1, we have \eqref{eqA2}:

\begin{equation}
\int_{- \infty}^{- MS}{\left| f^{- 1}\left( \mathbf{f}(s) \right) - s \right|P(s)d(s)} + \int_{MS}^{+ \infty}{\left| f^{- 1}\left( \mathbf{f}(s) \right) - s \right|P(s)d(s)} = 2\int_{MS}^{+ \infty}{(s - MS)P(s)d(s)}
\label{eqA2}
\end{equation}

Based on \eqref{eq3.1} and \eqref{eq3.2} , we have \eqref{eqA3} :

\begin{equation}
\int_{MS}^{MS}{\left| f^{- 1}\left( \mathbf{f}(s) \right) - s \right|P(s)d(s)} \leq \int_{MS}^{MS}{\frac{MS}{h}P(s)d(s)}
\label{eqA3}
\end{equation}

Then:
\begin{equation}
\begin{split}
&\int_{- \infty}^{+ \infty}{\left| f^{- 1}\left( \mathbf{f}(s) \right) - s \right|P(s)d(s)} = \int_{MS}^{MS}{\left| f^{- 1}\left( \mathbf{f}(s) \right) - s \right|P(s)d(s)} + \int_{- \infty}^{- MS}{\left| f^{- 1}\left( \mathbf{f}(s) \right) - s \right|P(s)d(s)} \\
&+ \int_{MS}^{+ \infty}{\left| f^{- 1}\left( \mathbf{f}(s) \right) - s \right|P(s)d(s)} \leq \int_{MS}^{MS}{\frac{MS}{h}P(s)d(s)} + 2\int_{MS}^{+ \infty}{(s - MS)P(s)d(s)} \\
&= MS\left( \frac{1}{h}\Phi(MS) - \frac{1}{h}\Phi( - MS) \right) + 2\int_{MS}^{+ \infty}{sP(s)ds} - 2MS\int_{MS}^{+ \infty}{P(s)d(s)} \\
&= MS\left( \frac{1}{h}\Phi(MS) - \frac{1}{h}\Phi( - MS) \right) + 2\int_{MS}^{+ \infty}{sP(s)ds} - 2MS\left( 1 - \Phi(MS) \right)\ \  \\
&= MS\left( \frac{1}{h}\left( \Phi(MS) - \Phi( - MS) \right) - 2 + 2\Phi(MS) \right) + 2\int_{MS}^{+ \infty}{sP(s)ds} \\
&= MS\left( \frac{1}{h}\left( \Phi(MS) - \Phi( - MS) \right) - 2 + 2\Phi(MS) \right) + \sqrt{\frac{2}{\pi}}e^{\frac{- MS^{2}}{2}}
\end{split}
\label{eqA4}
\end{equation}

This establishes \emph{\textbf{Theorem 1}}.

\vspace{-5pt}
\paragraph{Proposition 1}: When \(h \longrightarrow + \infty\),
\(\forall\varepsilon > 0,\ \exists\delta,\ \forall MS \in R^{+}\)and\(\ \forall MS \geq \delta\),
\eqref{eqP1} holds.

\begin{equation*}
\left| MS\left( \frac{1}{h}\left( \Phi(MS) - \Phi( - MS) \right) - 2 + 2\Phi(MS) \right) + \sqrt{\frac{2}{\pi}}e^{\frac{- MS^{2}}{2}} \right| \leq \varepsilon\ \\
\end{equation*}

\emph{Proof}:

When \(h \longrightarrow + \infty\), we have:

\[\]

\begin{equation}
\begin{split}
 \lim_{h \longrightarrow + \infty}\left| MS\left( \frac{1}{h}\left( \Phi(MS) - \Phi( - MS) \right) - 2 + 2\Phi(MS) \right) + \sqrt{\frac{2}{\pi}}e^{\frac{- MS^{2}}{2}} \right|\\
 = \left| MS\left( - 2 + 2\Phi(MS) \right) + \sqrt{\frac{2}{\pi}}e^{\frac{- MS^{2}}{2}} \right|
\label{eqA5}
\end{split}
\end{equation}

\begin{equation}
\begin{split}
 \lim_{MS \longrightarrow + \infty}\left| MS\left( - 2 + 2\Phi(MS) \right) \right| &= \lim_{MS \longrightarrow + \infty}\left| MS\left( - 2 + 2\Phi(MS) \right) \right| = \lim_{MS \longrightarrow + \infty}\left| \frac{{erf}{\left( \frac{MS}{\sqrt{2}} \right)\ } - 1}{\frac{1}{MS}} \right| \\
 &= \lim_{MS \longrightarrow + \infty}\left| - \frac{\sqrt{2}}{\sqrt{\pi}}{MS}^{2}e^{- \frac{{MS}^{2}}{2}} \right| = 0\
\label{eqA6}
\end{split}
\end{equation}

\begin{equation}
\lim_{MS \longrightarrow + \infty}\left| \sqrt{\frac{2}{\pi}}e^{\frac{- MS^{2}}{2}} \right| = 0
\label{eqA7}
\end{equation}

We have
\begin{equation}
\lim_{MS \longrightarrow + \infty}\left| MS\left( - 2 + 2\Phi(MS) \right) + \sqrt{\frac{2}{\pi}}e^{\frac{- MS^{2}}{2}} \right| = 0
\label{eqA8}
\end{equation}

This establishes \textbf{\emph{Proposition} \emph{1}}.

\vspace{-5pt}
\paragraph{Proposition 2}: Given h, there always exists a best \(MS^{*}\)
satisfied (6) to the minimize upper bound of SME.

\begin{equation*}
\frac{1}{h}\left( \Phi\left( MS^{*} \right) - \Phi\left( - MS^{*} \right) \right) - 2 + 2\Phi\left( MS^{*} \right) + \frac{MS^{*}}{h}\frac{\sqrt{2}}{\sqrt{\pi}}e^{- \frac{{MS^{*}}^{2}}{2}} = 0\
\end{equation*}

\emph{Proof}: Note the upper bound of SME as \(\chi(MS,h)\), we have:

\begin{equation}
\frac{\partial\chi(MS,h)}{\partial MS} = \frac{1}{h}\left( \Phi(MS) - \Phi( - MS) \right) - 2 + 2\Phi(MS) + \frac{MS}{h}\frac{\sqrt{2}}{\sqrt{\pi}}e^{- \frac{{MS}^{2}}{2}}
\label{eqA9}
\end{equation}
\begin{equation}
\frac{\partial^{2}\chi(MS,h)}{\partial^{2}MS} = \frac{\sqrt{2}}{\sqrt{\pi}}e^{- \frac{{MS}^{2}}{2}}\left( \frac{2 + h - MS^{2}}{h} \right)
\label{eqA10}
\end{equation}

Then we have:

\begin{equation}
\lim_{MS \rightarrow 0}\frac{\partial\chi(MS,h)}{\partial MS} < 0\ \ \ \ \ \lim_{MS \rightarrow + \infty}\frac{\partial\chi(MS,h)}{\partial MS} > 0
\label{eqA11}
\end{equation}

Considering the second order derivative of \(\chi(MS,h)\) will be
greater than 0 when \(MS < \sqrt{h + 2}\) and then lower than 0, and the
first 

\section{Visualizations}\label{appendix:illstruction}

\begin{figure}[htbp]
    \centering
    \subfigure[Forecasting result in \textit{V}]{
        \includegraphics[width=0.8\textwidth]{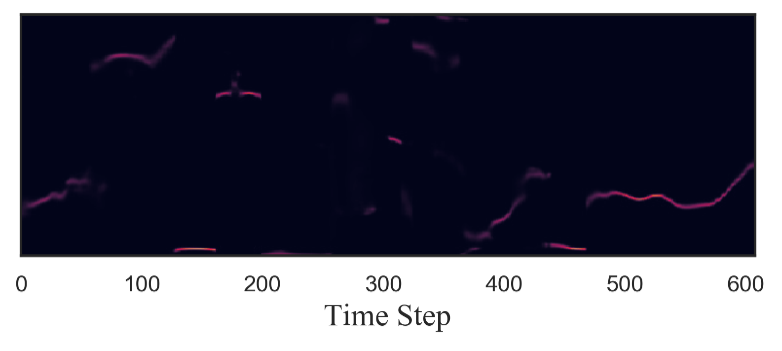}
    }
    \\
    \subfigure[Forecasting result in \textit{S}]{
        \includegraphics[width=0.8\textwidth]{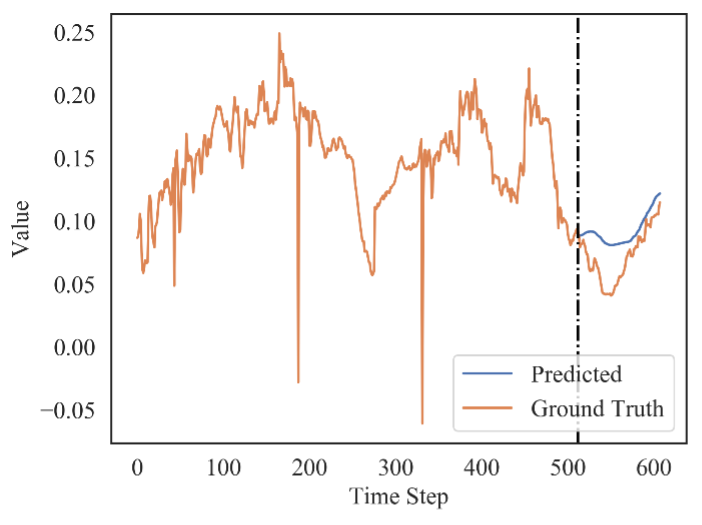}
    }
    \caption{The forecasting results of dataset Weather (prediction length =96).}
    \label{fA1}
\end{figure}

\begin{figure}[htbp]
    \centering
    \subfigure[Forecasting result in \textit{V}]{
        \includegraphics[width=0.8\textwidth]{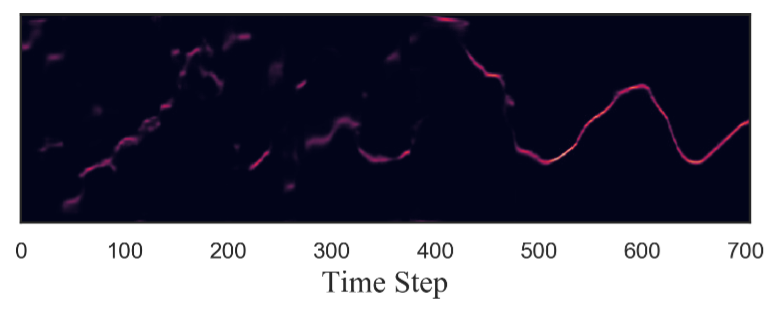}
    }
    \\
    \subfigure[Forecasting result in \textit{S}]{
        \includegraphics[width=0.8\textwidth]{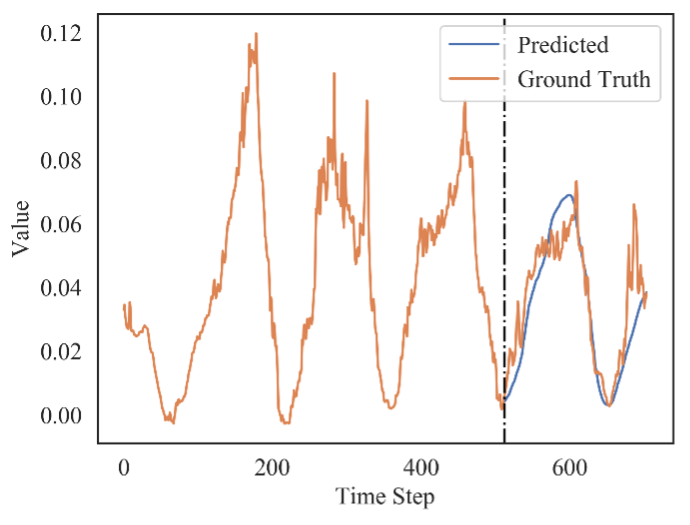}
    }
    \caption{The forecasting results of dataset Weather (prediction length =192).}
    \label{fA2}
\end{figure}

\begin{figure}[htbp]
    \centering
    \subfigure[Forecasting result in \textit{V}]{
        \includegraphics[width=0.8\textwidth]{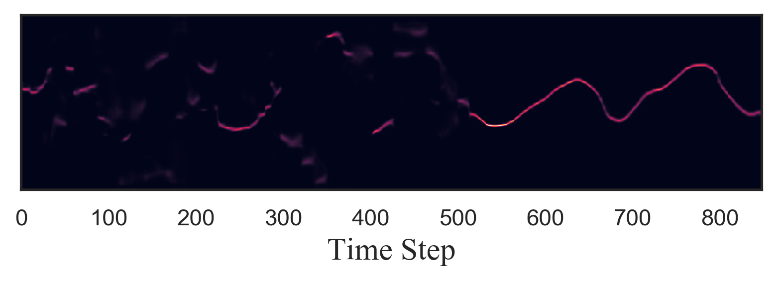}
    }
    \\
    \subfigure[Forecasting result in \textit{S}]{
        \includegraphics[width=0.8\textwidth]{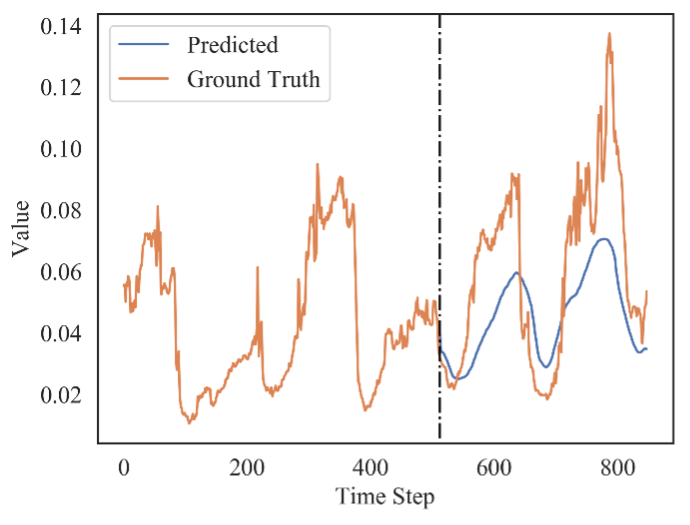}
    }
    \caption{The forecasting results of dataset Weather (prediction length =336).}
    \label{fA3}
\end{figure}

\begin{figure}[htbp]
    \centering
    \subfigure[Forecasting result in \textit{V}]{
        \includegraphics[width=0.8\textwidth]{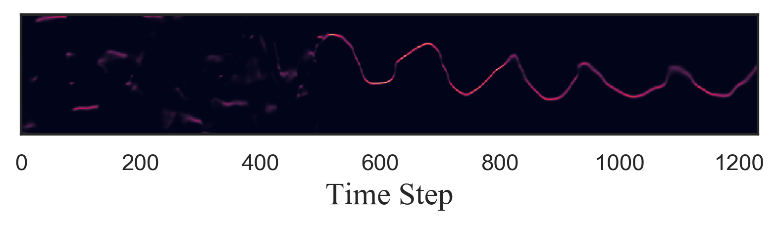}
    }
    \\
    \subfigure[Forecasting result in \textit{S}]{
        \includegraphics[width=0.8\textwidth]{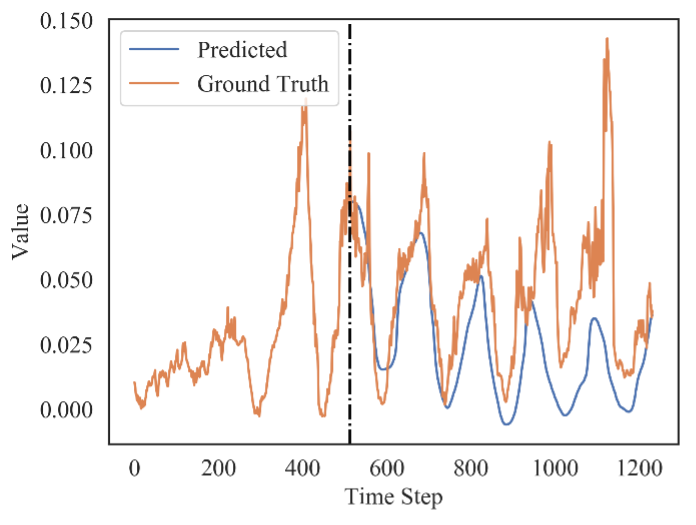}
    }
    \caption{The forecasting results of dataset Weather (prediction length =720).}
    \label{fA4}
\end{figure}

\begin{figure}[htbp]
    \centering
    \subfigure[Forecasting result in \textit{V}]{
        \includegraphics[width=0.3\textwidth]{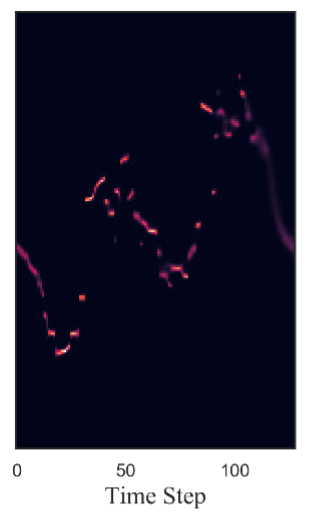}
    }
    \subfigure[Forecasting result in \textit{S}]{
        \includegraphics[width=0.6\textwidth]{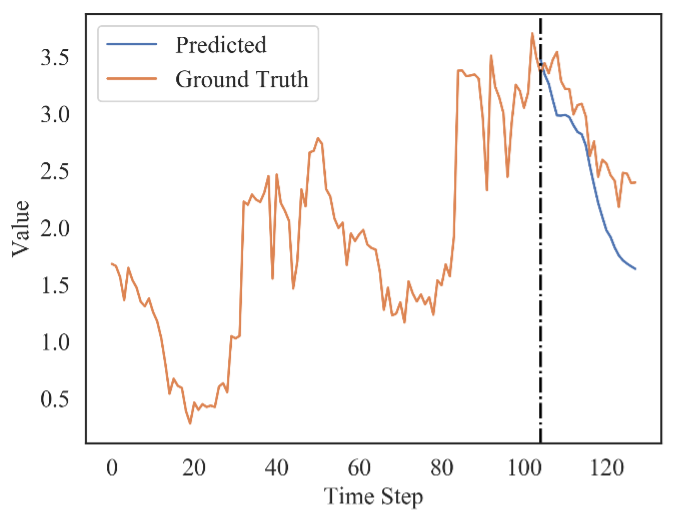}
    }
    \caption{The forecasting results of dataset ILI (prediction length =24).}
    \label{fA5}
\end{figure}

\begin{figure}[htbp]
    \centering
    \subfigure[Forecasting result in \textit{V}]{
        \includegraphics[width=0.3\textwidth]{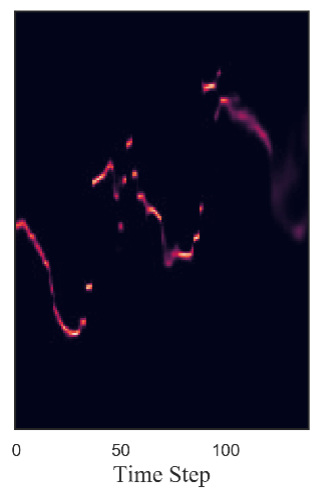}
    }
    \subfigure[Forecasting result in \textit{S}]{
        \includegraphics[width=0.6\textwidth]{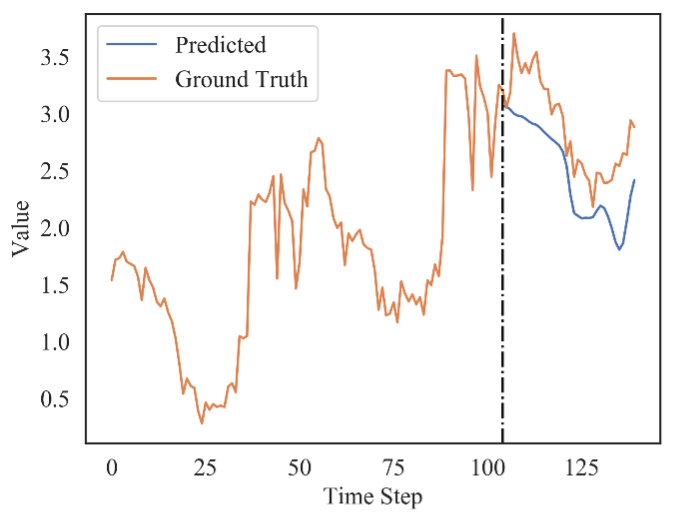}
    }
    \caption{The forecasting results of dataset ILI (prediction length =36).}
    \label{fA6}
\end{figure}

\begin{figure}[htbp]
    \centering
    \subfigure[Forecasting result in \textit{V}]{
        \includegraphics[width=0.3\textwidth]{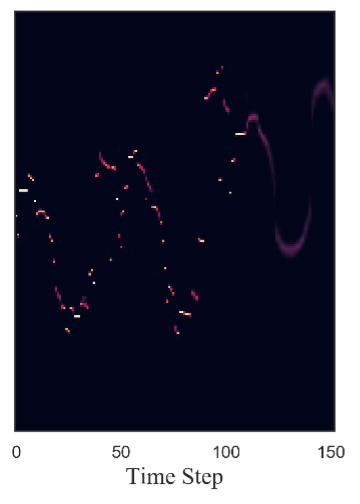}
    }
    \subfigure[Forecasting result in \textit{S}]{
        \includegraphics[width=0.6\textwidth]{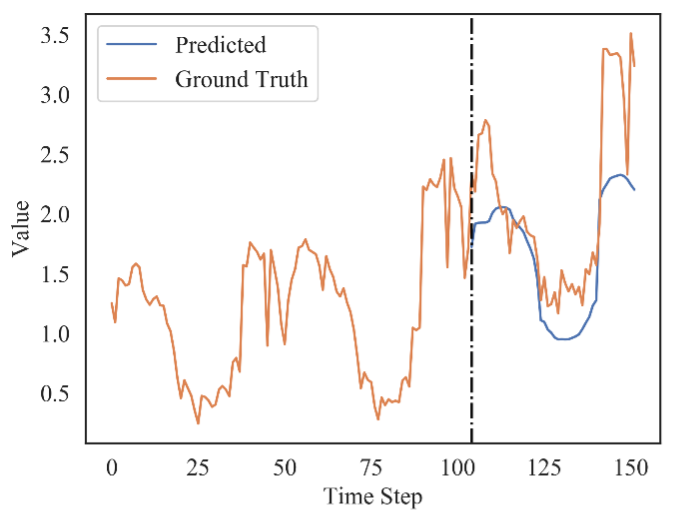}
    }
    \caption{The forecasting results of dataset ILI (prediction length =48).}
    \label{fA7}
\end{figure}

\begin{figure}[htbp]
    \centering
    \subfigure[Forecasting result in \textit{V}]{
        \includegraphics[width=0.3\textwidth]{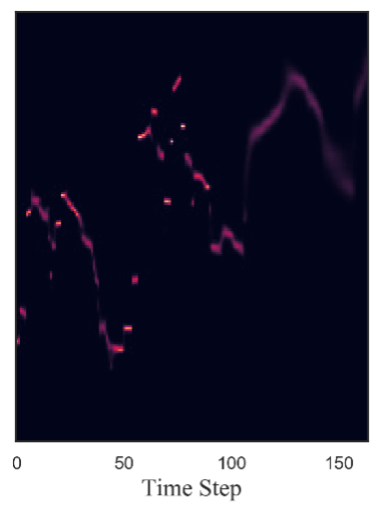}
    }
    \subfigure[Forecasting result in \textit{S}]{
        \includegraphics[width=0.6\textwidth]{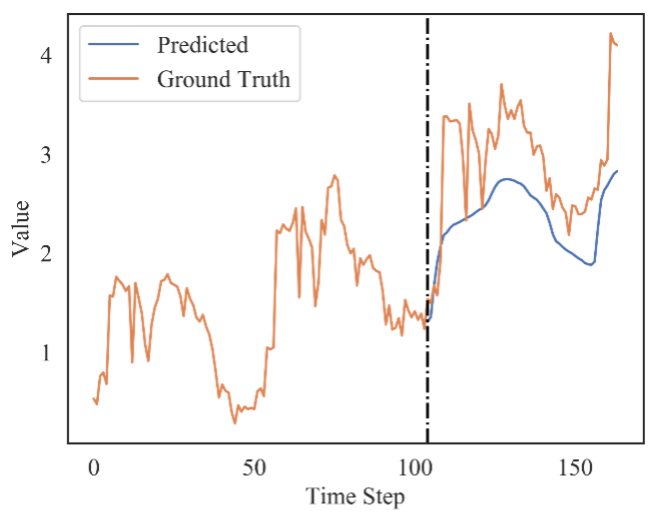}
    }
    \caption{The forecasting results of dataset ILI (prediction length =60).}
    \label{fA8}
\end{figure}
\end{document}